\documentclass{article}

\usepackage[preprint]{neurips_2025}

\usepackage{afterpage}

\usepackage{graphicx}
\usepackage{booktabs}
\usepackage{subcaption}
\usepackage{amsmath, amssymb, amsfonts}
\usepackage{multirow}
\usepackage{tabularx}
\usepackage{enumitem}
\usepackage{wrapfig}
\usepackage{float}
\usepackage{nicefrac}
\usepackage{microtype}
\usepackage[table,dvipsnames,x11names]{xcolor}
\usepackage{pifont}
\usepackage[most]{tcolorbox}
\usepackage{listings}
\usepackage{makecell}
\usepackage[font=small,tableposition=top]{caption}
\usepackage{hyperref}
\usepackage{cleveref}
\usepackage{orcidlink}
\definecolor{mydarkgreen}{HTML}{009901}
\definecolor{cvprblue}{rgb}{0.21,0.49,0.74}

\newcommand{\cmark}{\textcolor{teal}{\checkmark}}
\newcommand{\xmark}{\textcolor{red}{\ding{55}}}

\usepackage[export]{adjustbox}
\newcommand{\diffone}{\includegraphics[scale=0.15,valign=B]{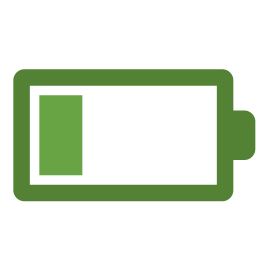}}
\newcommand{\difftwo}{\includegraphics[scale=0.15,valign=B]{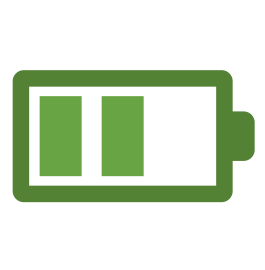}}
\newcommand{\diffthree}{\includegraphics[scale=0.15,valign=B]{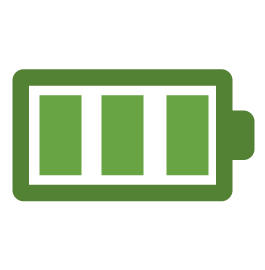}}

\newcommand\blfootnote[1]{%
  \begingroup
  \renewcommand\thefootnote{}\footnote{#1}%
  \addtocounter{footnote}{-1}%
  \endgroup
}

\makeatletter
\def\blfootnote{\gdef\@thefnmark{}\@footnotetext}
\makeatother

\title{PerceptionComp: A Video Benchmark for Complex Perception-Centric Reasoning}

\author{
Shaoxuan Li$^{1 *}$ \quad
Zhixuan Zhao$^{1 *}$ \quad
Hanze Deng$^{1 *}$ \quad
Zirun Ma$^{1 *}$ \\
\textbf{Shulin Tian$^{3}$} \quad
\textbf{Zuyan Liu$^{1}$} \quad
\textbf{Yushi Hu$^{2}$} \quad
\textbf{Haoning Wu$^{3}$} \\
\textbf{Yuhao Dong$^{3 \#}$} \quad
\textbf{Benlin Liu$^{2 \#}$} \quad
\textbf{Ziwei Liu$^{3 \dagger}$} \quad
\textbf{Ranjay Krishna$^{2 \dagger}$} \\
\\
$^{1}$Tsinghua University \quad
$^{2}$University of Washington \quad
$^{3}$Nanyang Technological University
\\
}

\begin{document}

\maketitle

\blfootnote{$^{*}$ Equal contribution \quad $^{\#}$ Project co-lead \quad $^{\dagger}$ Equal advising}


\begin{abstract}
Deep video understanding requires long-horizon, perception-centric reasoning that repeatedly revisits a video to gather temporally distributed evidence. However, existing benchmarks are either relatively easy (perception-centric but often solvable after a single view) or logic-heavy with simplified visuals, and thus do not faithfully measure multimodal test-time thinking that depends on repeated perception.
We introduce \textbf{PerceptionComp}, a fully manually annotated benchmark designed so that no single moment is sufficient: answering requires evidence from multiple temporally separated segments under compositional constraints. PerceptionComp contains \textbf{1,114} five-choice questions over \textbf{279} high-scene-complexity videos spanning diverse domains. Videos are selected using automatic proxies for scene complexity (SAM2 instance counts and optical-flow magnitude), and each question requires \textbf{10--20 minutes} of annotation.
Human evaluation confirms the intended difficulty: PerceptionComp requires substantially longer response times than prior benchmarks, and under a single-view setting (no rewatching) human accuracy drops to near chance (\textbf{18.97\%}), while experts can reach \textbf{100\%} accuracy with unrestricted rewatching and sufficient time.
State-of-the-art MLLMs perform notably worse: the best model in our evaluation (Gemini-3-Flash) reaches only \textbf{45.96\%} accuracy, and open-source MLLMs remain below \textbf{40\%}. Test-time reasoning helps but remains far from human-level (e.g., GPT-o3 exceeds GPT-4o by \textbf{11.04\%}; Gemini-2.5-Pro exceeds Gemini-2.5-Flash by \textbf{6.19\%}), and increasing test-time compute via larger thinking-token budgets or more input frames further improves performance. Finally, among the strongest frontier models we tested (Gemini-3 variants and GPT-o3), accuracies cluster in the mid-40s, suggesting a bottleneck in perception-centric long-horizon video reasoning. PerceptionComp provides a focused testbed for diagnosing these limitations and advancing multimodal visual thinking.
\end{abstract}
\begin{figure}[t]
\centering
\includegraphics[width=\linewidth]{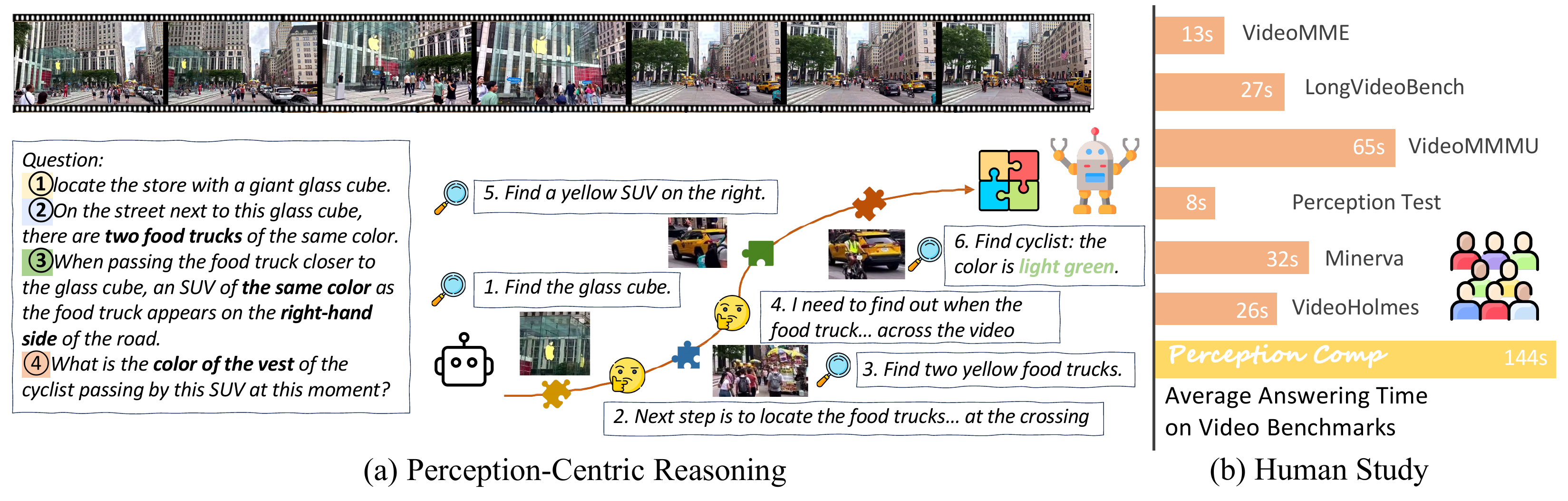}
\caption{\textbf{Overview of the PerceptionComp benchmark.}
(a) An example from PerceptionComp, where models are required to perform complex, perception-centric reasoning with various types of subconditions to arrive at the final answer.
(b) Results from a human study measuring question-answering time, showing that PerceptionComp is more challenging for humans than previous perception and reasoning video benchmarks, largely due to its emphasis on perception-centric reasoning.}
\label{fig:teaser}
\end{figure}

\section{Introduction}

Videos capture human activities and the physical world, and multimodal intelligence---from robots to AI glasses---must achieve \emph{deep} video understanding to be broadly useful. Consider a seemingly simple query: \emph{``On which floor did the person last appear in the video before dropping their \textbf{apartment} keys (not their office keys)?''} Answering it requires long-horizon, step-by-step repeated perception that composes multiple perceptual skills: \textit{semantic recognition} (identify which object is a key), \textit{correspondence} (track the apartment key rather than the office key), \textit{temporal reasoning} (locate the drop event and trace back to the previous time the key appears), and \textit{spatial reasoning} (infer the building layout and floor level). Recent breakthroughs in long-horizon reasoning for mathematics and coding suggest that \emph{test-time scaling}---allocating more computation during inference---is a promising route for enabling multimodal language models (MLLMs) to perform such multi-step, perception-driven video reasoning~\cite{guo2025deepseek,huang2025vision,videochatr1,videor1}. For deep video understanding, this should not mean only longer language-side thinking; it should also mean composing multiple perception skills and repeatedly revisiting the video to gather visual information across different dimensions.

However, existing video benchmarks do not adequately measure this capability. Many widely used benchmarks (e.g., VideoMME, Perception Test)~\cite{fu2024video,hu2025video,cheng2025video,patraucean2023perception} are perception-centric but relatively easy: humans can often answer after a single viewing with minimal deliberation (Fig.~\ref{fig:teaser}), leaving limited room to differentiate models' test-time thinking ability. In contrast, benchmarks that demand substantial reasoning, such as geometry or maze solving~\cite{rasheed2025videomathqa,wu2024vsp}, often derive difficulty primarily from logical structure rather than real-world perception, since their visual inputs are synthetic or overly simple. Even long-video understanding benchmarks~\cite{wu2024longvideobench} frequently stress memory more than evidence-seeking reasoning. As a result, there is still no benchmark that is simultaneously long-horizon, perception-centric, and truly forces repeated visual information gathering.

To fill this gap, we introduce \textbf{PerceptionComp}, a manually annotated benchmark for \emph{complex, compositional, and comprehensive perception-centric} video reasoning over long horizons. PerceptionComp is constructed so that no single moment is sufficient: solving a question requires multiple temporally separated pieces of visual evidence and compositional constraints. Concretely, each question combines several perceptual sub-conditions under two logics---\emph{conjunctive} and \emph{sequential}---so the model must satisfy multiple constraints and track entities/states across time. Each sub-condition is itself a perceptual subtask that requires extracting diverse visually grounded information from the video, including objects, attributes, relations, locations, actions, and events. Completing these subtasks requires a range of perceptual skills, including semantic recognition, visual correspondence, temporal reasoning, and spatial reasoning; some questions additionally involve commonsense knowledge tightly linked to the visual content and simple near-future prediction from ongoing dynamics.

We manually annotate PerceptionComp following this question design on 279 videos drawn from diverse domains, such as city walk tours, large indoor villa tours, video games, and extreme outdoor sports, resulting in 1,114 highly complex questions. We quantify scene complexity using automatic signals: specifically, we use the number of instances detected by SAM2~\cite{ravi2024sam} and optical-flow magnitude~\cite{teed2020raft} as proxies for object density, motion intensity, and scene-change dynamics, and we select high-complexity videos with many objects, intense motion, and frequent transitions. Beyond video complexity, each question has high compositional complexity and requires multiple perceptual reasoning skills to extract different aspects of visual evidence. To ensure correctness under this extreme difficulty, we adopt 100\% manual annotation: each question takes 10--20 minutes from video selection to final annotation. We use a five-choice format for reliable evaluation and report accuracy as the metric; answer options are designed to be plausible and closely confusable, so disambiguation requires video evidence rather than option priors.

We first evaluate PerceptionComp with humans and treat human performance as a gold standard for benchmark quality. In our human study, participants watch the video and answer; during answering, they may rewatch the video as needed. We measure response time and find that participants take substantially longer on PerceptionComp than on prior benchmarks (Fig.~\ref{fig:teaser}): more than $2\times$ longer than VideoMMMU~\cite{hu2025video}, more than $10\times$ longer than VideoMME~\cite{fu2024video}, more than $5\times$ longer than Video-Holmes~\cite{cheng2025video}, and more than $5\times$ longer than LongVideoBench~\cite{wu2024longvideobench}, even though most of our videos are not longer than 10 minutes. This shows that video context length is not the only dimension of video thinking. We further evaluate a stricter setting where participants may watch the video only once and cannot rewatch while answering; accuracy in this setting is near chance (18.97\%), even with extended thinking time. Together, these results show that PerceptionComp (i) requires substantial test-time thinking to solve and (ii) cannot be solved without repeated perception steps, ruling out shortcuts based on single-view memory or language priors. Given its coverage of diverse perceptual skills, PerceptionComp also serves as a testbed for perceptual competence.

We further evaluate PerceptionComp on state-of-the-art MLLMs. While these models achieve strong results on existing benchmarks, they perform notably worse on PerceptionComp. Even the best-performing model in our evaluation (Gemini-3-Flash~\cite{gemini2.5}) reaches only 45.96\% accuracy in the five-choice setting, and open-source MLLMs~\cite{qwen2.5vl,qwen3vl,internvl3,kimivl,mimovl} remain below 40\%. In contrast, human participants can reach 100\% accuracy when given sufficient time with unrestricted rewatching.
Moreover, test-time reasoning helps, but performance remains far from human-level. Thinking models outperform their non-thinking counterparts: GPT-o3~\cite{jaech2024openai} surpasses GPT-4o~\cite{gpt4o} by 11.04\%, and Gemini-2.5-Pro exceeds Gemini-2.5-Flash by 6.19\%. By controlling the thinking-token budget for Gemini-2.5-Flash~\cite{gemini2.5}, we further show that allocating more test-time tokens improves performance. We also find that providing more input frames boosts accuracy for both GPT-o3~\cite{jaech2024openai} and Qwen3-VL-8B~\cite{qwen3}, consistent with the benchmark’s reliance on temporally distributed evidence.
Finally, among the strongest frontier models we tested (Gemini-3 variants and GPT-o3), despite different architectures/interfaces, accuracies cluster in the mid-40s, suggesting a potential bottleneck in perception-centric long-horizon video reasoning. To better understand this regime, we analyze representative failure cases of frontier models and characterize common bottlenecks. We hope PerceptionComp will help the community recognize these limitations and drive progress in perceptual reasoning.

\begin{table}[t]
  \centering
  \small 
  
  \caption{\textbf{Comparison of PerceptionComp with other benchmarks.} PerceptionComp distinguishes itself from previous benchmarks by emphasizing perception-centric reasoning, assessing how models integrate visual evidence with reasoning processes.}
  \label{tab:compare_benchmark}
  \vspace{8pt}
  
  \setlength\tabcolsep{6pt} 
  \begin{tabular}{@{}l l c c c@{}}
    \toprule
    \multirow{2}{*}{\textbf{Benchmark}} & \multicolumn{4}{c}{\textbf{Properties}} \\
    \cmidrule(lr){2-5}
    & \textbf{Video Domain} & \textbf{\# QA} & \textbf{Per.Rea} & \textbf{Annotation} \\
    \midrule
    MMVU~\cite{zhao2025mmvu}           & Educational videos  & 3,000 & \xmark & Manual    \\
    VideoMME~\cite{fu2024video}         & YouTube videos      & 2,700 & \xmark & Manual    \\
    VCR-Bench~\cite{qi2025vcr}          & Short films         & 1,034 & \xmark & Manual    \\
    MINERVA~\cite{nagrani2025minerva}   & Mix                 & 1,515 & \xmark & Manual    \\
    VideoMMMU~\cite{hu2025video}        & Lectures            & 900   & \xmark & Manual    \\
    Video-Holmes~\cite{cheng2025video}  & Short films         & 1,834 & \xmark & Automatic\&Manual \\
    \midrule
    \rowcolor{gray!10} 
    \textbf{PerceptionComp} & \textbf{In-the-wild videos} & \textbf{1114} & \cmark & \textbf{Manual} \\
    \bottomrule
  \end{tabular}
\end{table}

\section{Related Work}
\label{sec:formatting}

\noindent \textbf{General Video Understanding Benchmarks.}
Traditional video understanding benchmarks often focus on relatively basic perceptual understanding---either local details (e.g., short clips or fine-grained actions) or global summaries---with outcome-based metrics. Recent general-purpose benchmarks like Video-MME~\cite{fu2024video} and ALLVB~\cite{tan2025allvb} broaden task coverage across domains and video lengths, while task-specific suites such as MVBench~\cite{li2024mvbench} and NExT-QA~\cite{xiao2021next} isolate skills like temporal reasoning and object interaction. The Perception Test~\cite{patraucean2023perceptiontest} further provides diagnostic, perception-oriented evaluation on purposefully designed real-world videos. However, these benchmarks are typically solvable with limited cross-moment evidence integration and thus remain comparatively \emph{easy} as probes of long-horizon, compositional video thinking. Long-video benchmarks~\cite{wang2025lvbench,wu2024longvideobench,rawal2024cinepile,song2024moviechat} emphasize memory and narrative comprehension over extended durations but largely reduce evaluation to single-turn QA. Egocentric benchmarks~\cite{grauman2022ego4d,mangalam2023egoschema} add realism through first-person perspectives, but they typically do not require repeated perception to iteratively gather diverse visual evidence across multiple segments. PerceptionComp differs by making difficulty \emph{perception-bottlenecked} through long-horizon compositional queries that require repeated evidence gathering.

\noindent \textbf{Complex Multimodal Reasoning Benchmarks.}
Recent progress in multimodal reasoning has led to benchmarks that go beyond surface-level understanding to evaluate structured inference across vision and language. In the image domain, VCBench and related benchmarks~\cite{li2024vcbench,hao2025can,xu2025visulogic} target mathematical, scientific, and logical reasoning where visual inputs mainly serve as a carrier of symbolic structure, and ScienceQA~\cite{saikh2022scienceqa} and EXAMS-V~\cite{das2024exams} introduce academic-style questions that emphasize explanation and cross-domain knowledge. In video, early benchmarks~\cite{xu2017video,yu2019activitynet,xiao2021next} focus on short-term understanding, while later ones~\cite{li2024mvbench,liu2024mmbench,liu2024tempcompass} add richer temporal structure but often remain relatively shallow. Long-context benchmarks~\cite{wu2024longvideobench,fu2024video} scale to longer videos, yet many questions can still be answered from isolated cues. More advanced evaluations~\cite{zhao2025mmvu,hu2025video,rasheed2025videomathqa,yang2025thinking} target scientific, academic, or spatial understanding, and VCR-Bench~\cite{qi2025vcr} and MME-CoT~\cite{jiang2025mme} begin to assess chain-of-thought behavior. Recent benchmarks such as MINERVA~\cite{nagrani2025minerva} and Video-Holmes~\cite{cheng2025video} further emphasize multi-step temporal and causal inference. However, across many of these ``hard'' settings, difficulty is often dominated by \emph{logical} or \emph{domain} reasoning (e.g., math/science/geometry), with comparatively lightweight perceptual demands. In contrast (Table~\ref{tab:compare_benchmark}), PerceptionComp makes the bottleneck \emph{perception}: questions are designed so that no single moment is sufficient and solving them requires repeatedly gathering fine-grained visual evidence across temporally separated segments, yielding a more faithful probe of perception-centric compositional video reasoning.

\noindent \textbf{Multimodal Reasoning Models.}
Reasoning-oriented LLMs show that long-horizon inference benefits from step-by-step reasoning and test-time scaling. In parallel, MLLMs have evolved from early image--text systems to unified models that directly accept visual inputs. Frontier proprietary models (GPT-style, Gemini-style)~\cite{gpt4o,gemini2.5} and strong open-source families (Qwen-VL, InternVL, Molmo)~\cite{qwen2.5vl,qwen3,internvl3,clark2026molmo2} support direct image understanding, and increasingly extend the same interface to videos for end-to-end video QA.
Following DeepSeek-R1~\cite{guo2025deepseek} and the shift toward long-form reasoning, recent work begins to bring RLVR-style or reasoning-focused pipelines to multimodal models. In images, Vision-R1/VisualRFT~\cite{huang2025vision,liu2025visual} and DeepEyes~\cite{Zheng2025DeepEyesI} improve visual reasoning with verifiable rewards or interleaved multimodal traces. In videos, Video-R1~\cite{videor1} and VideoChat-R1~\cite{videochatr1} elicit longer reasoning traces for multi-step temporal inference. Our benchmark complements these efforts by providing a perception-centric, long-horizon testbed that stresses repeated evidence gathering, where difficulty is dominated by perception rather than purely logical structure.        
\vspace{-2pt}
\section{PerceptionComp}

\begin{figure}[t]
    \centering
    \includegraphics[width=\linewidth]{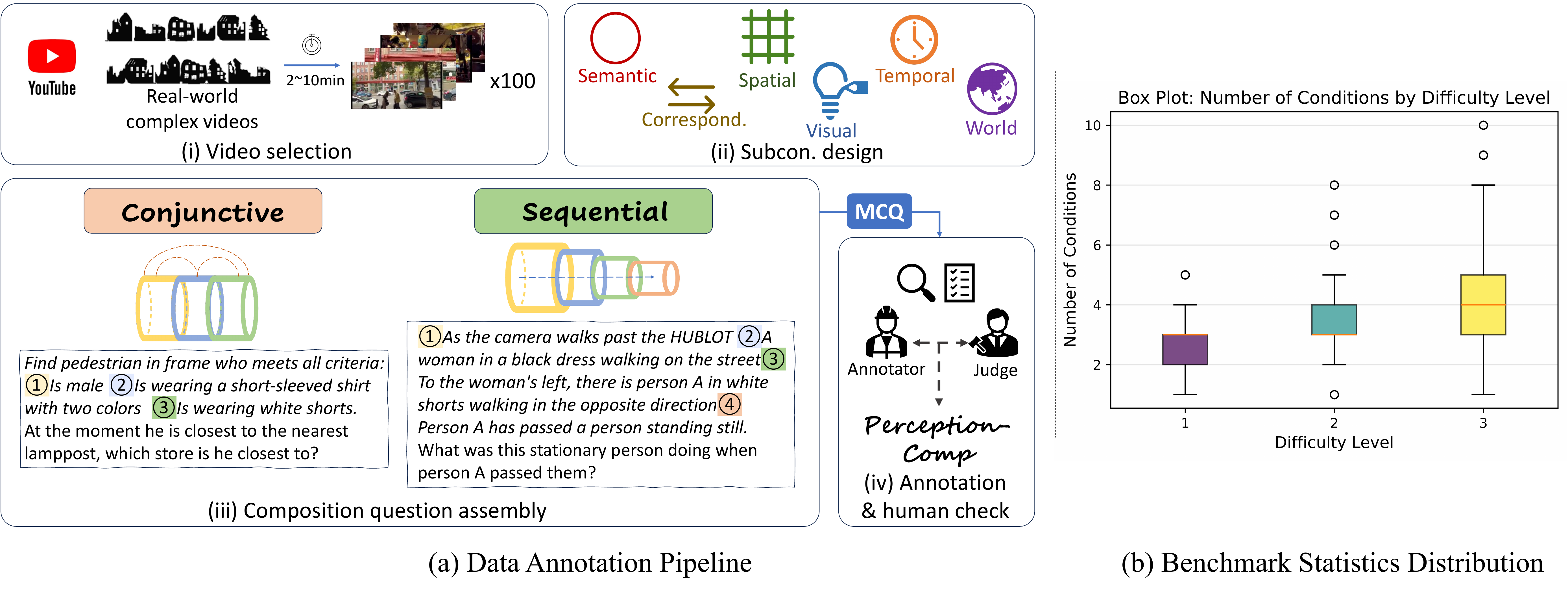}
    \caption{\textbf{Data construction and statistics of PerceptionComp.} (a) Annotation pipeline, which integrates diverse subconditions and supports two types of compositional questions. (b) Benchmark statistics: higher difficulty levels contain more subconditions, increasing the demand for perception-centric reasoning.}
    \label{fig:bench_stat}
\end{figure}

\begin{figure}[t]
    \centering
    \includegraphics[width=0.95\linewidth]{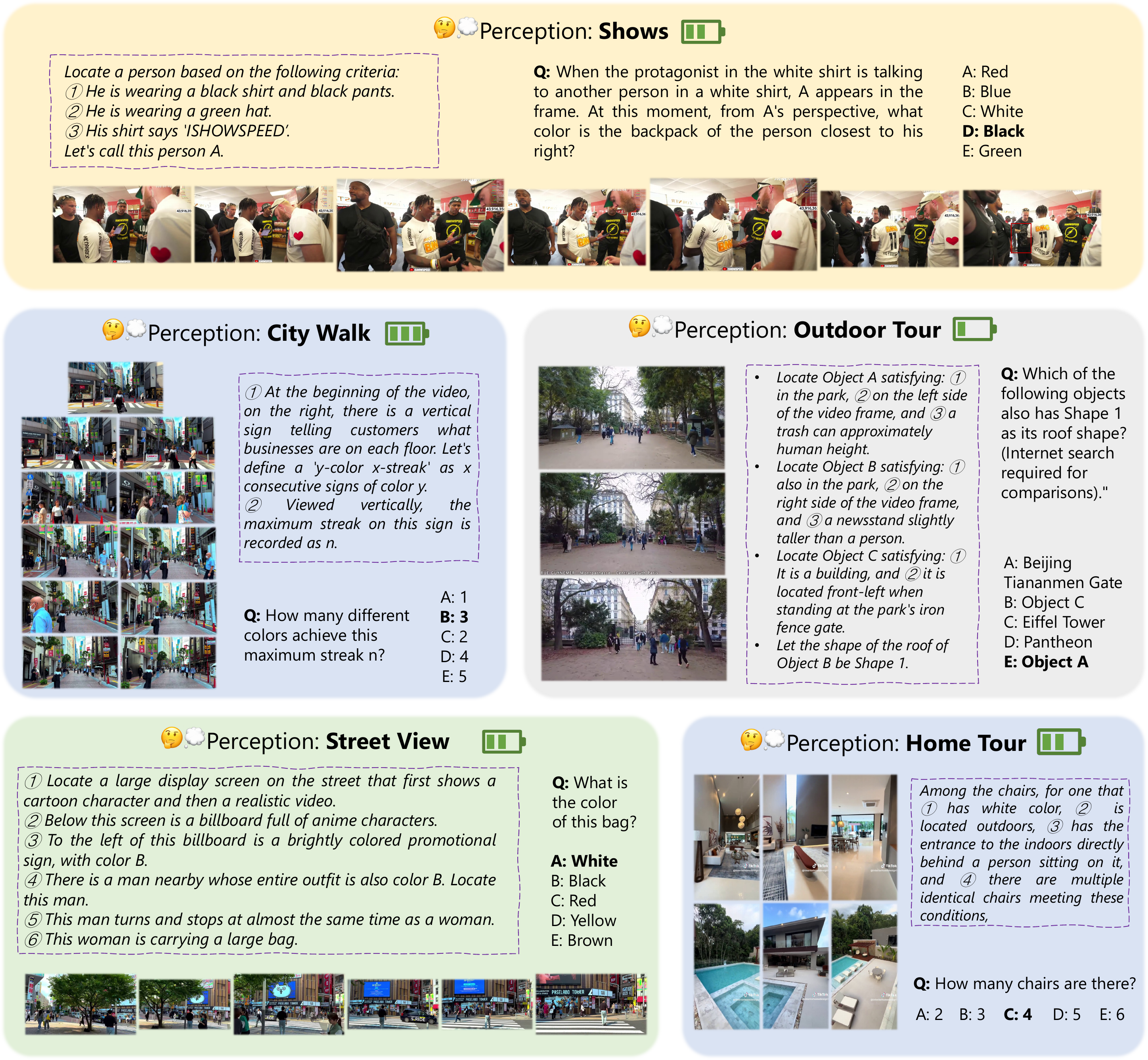}
    \caption{\textbf{Examples from PerceptionComp.} \diffone{}, \difftwo{}, and \diffthree{} denote difficulty levels 1, 2, and 3, respectively. PerceptionComp spans diverse video sources and uses subconditions to construct conjunctive and sequential questions that require perception-centric reasoning.}
    \label{fig:bench_examples}
\end{figure}

We design PerceptionComp to evaluate video thinking by enforcing long-horizon, perception-centric reasoning. We do so in two ways: (i) selecting structurally complex videos, and (ii) composing questions from multiple subconditions that probe different perceptual skills, thereby increasing compositional complexity. Fig.~\ref{fig:bench_stat}(a) summarizes our end-to-end manual annotation pipeline, and Fig.~\ref{fig:bench_examples} provides representative question examples across difficulty levels and composition types. Below, we describe our video selection process, the question--answer format, the annotation pipeline, and our difficulty annotation.

\subsection{Video Selection}
Many existing video benchmarks use clips that are visually simple: they often depict a single event or activity and contain only a small number of humans or objects. As a result, many videos can be approximately replaced by a short textual caption without substantially affecting downstream performance, limiting their ability to diagnose perceptual competence.

To better probe perception-related abilities, we deliberately select videos with high scene and object complexity. Our videos span seven categories: city-walk tours (outdoor), shopping in malls, sports competitions, indoor home/villa tours, variety shows, movie clips, and game livestreams. These videos typically contain many objects, frequent scene transitions, and substantial camera motion, making them difficult to summarize with a single caption. The selected clips range from 2 to 10 minutes in length. Unlike benchmarks that increase difficulty primarily by extending duration, we also increase difficulty along an orthogonal axis: dynamic scene complexity.

We quantify complexity using automatic signals: we use the number of instances detected by SAM2~\cite{ravi2024sam} and optical-flow magnitude~\cite{teed2020raft} as proxies for object density, motion intensity, and scene-change dynamics, and prioritize clips with many objects, intense motion, and frequent transitions. All videos are sourced from real recordings rather than synthetic renderings; while some categories (e.g., game livestreams) are screen-captured, the videos still exhibit rich, naturally occurring dynamics and clutter that make the tasks challenging and practically relevant. This combination of dynamic scene complexity and rapidly changing content forces models to repeatedly gather and integrate visual evidence across temporally separated moments rather than relying on a coarse global summary. We defer more detailed video statistics to the supplementary.

\subsection{Subconditions and Perceptual Skills}
We explicitly increase compositional difficulty by combining multiple subconditions into a single query. Each subcondition targets a distinct perceptual--reasoning skill, so solving the full question requires coordinated use of multiple abilities rather than a single narrow competence. Concretely, our subconditions cover:
\begin{itemize}
    \item \textbf{Semantic understanding}: Recognize object categories, attributes (e.g., shape, color, material), and higher-level semantic relations (e.g., roles or interactions).
    \item \textbf{Spatial understanding}: Reason about scene layout and relative geometry (e.g., left/right, front/behind, near/far) and occlusion.
    \item \textbf{Temporal understanding}: Follow motion patterns and localize events in time (e.g., what happens before/after a reference event).
    \item \textbf{Correspondence}: Match instances or parts across time/views (e.g., tracking the same object across shots, or part--whole matching).
    \item \textbf{Visual knowledge}: Some questions require commonsense knowledge tightly coupled to visual content.
    \item \textbf{World modeling}: Some questions involve simple near-future prediction from ongoing dynamics.
\end{itemize}
By sampling and composing subconditions from these categories, each question assesses perceptual competence in complex video reasoning more comprehensively than single-skill probes. Example subconditions and their compositions are illustrated in Fig.~\ref{fig:bench_examples}.

\subsection{Compositional Question Design}
We combine subconditions into full questions using two composition logics. 
\noindent \textbf{Conjunctive composition:} All subconditions refer to the \emph{same} target, forming an ``and''-style conjunction. To ensure every subcondition matters, we verify that no proper subset uniquely determines the answer: each condition only removes part of the candidate set, and only the full conjunction yields a single solution. This prevents shortcuts where the model can ignore part of the query. 

\noindent \textbf{Sequential composition:} Subconditions must be resolved \emph{in order}, where later subconditions depend on intermediate entities or states established earlier. For example, a first subcondition identifies an object, the second constrains its behavior at a later time, and a third asks about a relation involving that same object after another event. The model must carry the referent forward across steps, inducing a multi-hop perceptual reasoning process in which early errors propagate.

\subsection{Answer Space}
Each question is formed by composing subconditions, and the final answer is a piece of perceptual information extracted from the video. Answers fall into six categories:
\begin{itemize}
    \item \textbf{Objects}: Category names (e.g., ``car'', ``sofa'').
    \item \textbf{Attributes}: Properties such as color, count, or shape.
    \item \textbf{Relationships}: Semantic/spatial/social relations between entities.
    \item \textbf{Location}: Place descriptors (e.g., room type, country, or region).
    \item \textbf{Action}: The name of an action performed by an agent.
    \item \textbf{Event}: A higher-level event or composite situation that occurs in the video.
\end{itemize}
We cast every question as a five-way multiple-choice problem. To discourage reliance on language priors, all distractors are drawn from the \emph{same} answer category as the correct option (e.g., all colors or all object categories). Each option is constrained to a single word or a very short phrase, minimizing extra linguistic cues.

\subsection{Difficulty Annotation}
To enable difficulty-aware evaluation, we ask expert annotators to assign each question to one of three difficulty levels (Level 1/2/3) based on both (i) the number of composed subconditions and (ii) the intrinsic difficulty of the subconditions. This avoids treating difficulty as a function of subcondition count alone. As shown in Fig.~\ref{fig:bench_stat}(b), higher difficulty levels typically contain more subconditions, reflecting increased compositional depth and stronger requirements for long-horizon, perception-centric reasoning.

\subsection{Annotation Pipeline}
Following the procedure above, we select 279 videos with high scene complexity and annotate 1,114 questions. Because each question is highly compositional, we adopt fully manual annotation to ensure correctness. Representative annotated examples are shown in Fig.~\ref{fig:bench_examples}. Annotators first create the subconditions and final answer, then verify that (i) the answer is uniquely determined by the video and (ii) every subcondition is necessary.

Each question is subsequently checked by at least one additional annotator who did not create it. During verification, we confirm again that there is a single correct answer and that no proper subset of subconditions suffices to uniquely identify it. Items that fail either requirement are revised or discarded. This protocol ensures each question admits a unique solution and genuinely requires the full set of composed perceptual subconditions.

\definecolor{highlightgreen}{RGB}{235, 255, 235}
\definecolor{highlightred}{RGB}{255, 235, 235}

\begin{table}[!t]
\centering
\caption{\textbf{Comprehensive evaluation results of MLLMs on PerceptionComp.} We report both category-wise accuracies and accuracies across different difficulty levels.}
\vspace{8pt}
\label{tab:overall_results_final}
\small
\setlength{\tabcolsep}{3.5pt}
\begin{adjustbox}{max width=\textwidth}
\begin{tabular}{@{} l cc ccccccc ccc c @{}}
\toprule
\multirow{2.5}{*}{\textbf{Model}} & \multirow{2.5}{*}{\textbf{Size}} & \multirow{2.5}{*}{\textbf{Frame}} & \multicolumn{7}{c}{\textbf{Accuracy by Category}} & \multicolumn{3}{c}{\textbf{Accuracy by Difficulty}} & \multirow{2.5}{*}{\textbf{Overall}} \\
\cmidrule(lr){4-10} \cmidrule(lr){11-13}
& & & \rotatebox{45}{Outdoor} & \rotatebox{45}{Shopping} & \rotatebox{45}{Sport} & \rotatebox{45}{Home} & \rotatebox{45}{Show} & \rotatebox{45}{Movie} & \rotatebox{45}{Game} & \rotatebox{45}{Level 1} & \rotatebox{45}{Level 2} & \rotatebox{45}{Level 3} & \\
\midrule

\rowcolor{gray!5}

\multicolumn{14}{l}{\textit{\textbf{Human Performance}}} \\
Expert (unrestricted rewatch) & - & - & 100.00 & 100.00 & 100.00 & 100.00 & 100.00 & 100.00 & 100.00 & 100.00 & 100.00 & 100.00 & \textbf{100.00} \\
Human & - & - & 81.33 & 86.80 & 87.56 & 86.72 & 87.25 & 88.00 & 87.10 & 90.18 & 86.00 & 72.25 & \textbf{85.10} \\
Single-view Human (no rewatch) & - & - & 19.40 & 17.80 & 20.60 & 18.20 & 21.10 & 17.00 & 18.90 & 20.30 & 19.10 & 17.60 & \textbf{18.97} \\

\midrule
\multicolumn{14}{l}{\textit{\textbf{Proprietary Models}}} \\
Gemini-3-Flash~\cite{gemini2.5} & -   & - & 45.27 & 45.18 & 38.34 & 42.97 & 59.73 & 52.00 & 48.39 & 43.75 & 47.26 & 47.85 & 45.96 \\
Gemini-3-Pro~\cite{gemini2.5} & -   & - & 42.20 & 42.13 & 40.41 & 37.50 & 60.40 & 48.00 & 61.29 & 45.09 & 45.51 & 40.67 & 44.43 \\
Gemini-2.5-Pro~\cite{gemini2.5}   & -   & - & 45.78 & 45.18 & 32.64 & 42.19 & 52.35 & 52.00 & 58.06 & 44.64 & 44.20 & 44.02 & 44.34 \\
Seed-2.0-Pro~\cite{doubaovl}     & -     & 64 & 43.73 & 47.72 & 30.57 & 47.66 & 51.68 & 40.00 & 70.97 & 48.21 & 45.51 & 33.49 & 44.34 \\
Gemini-3.1-Pro~\cite{gemini2.5} & -   & - & 39.90 & 44.16 & 39.90 & 42.97 & 56.38 & 48.00 & 51.61 & 45.76 & 45.08 & 36.36 & 43.72 \\
GPT-o3~\cite{jaech2024openai} & - & 50 & 43.22 & 46.70 & 32.64 & 44.53 & 50.34 & 56.00 & 48.39 & 43.08 & 43.98 & 43.54 & 43.54 \\
GPT-5.2~\cite{gpt4} & - & 64 & 42.97 & 44.16 & 27.46 & 38.28 & 48.99 & 48.00 & 38.71 & 44.42 & 38.07 & 38.76 & 40.75 \\
Gemini-2.5-Flash~\cite{gemini2.5} & -   & - & 39.39 & 41.12 & 24.35 & 39.84 & 47.65 & 44.00 & 32.26 & 41.96 & 35.89 & 34.93 & 38.15 \\
GPT-5~\cite{gpt4}    & -   & 64  & 26.60 & 47.21 & 29.53 & 40.62 & 49.66 & 56.00 & 38.71 & 40.18 & 36.32 & 28.71 & 36.45 \\
GPT-4.1~\cite{gpt4} & - & 50 & 26.09 & 46.70 & 27.46 & 28.91 & 44.97 & 52.00 & 48.39 & 37.28 & 34.79 & 24.40 & 34.02 \\ 
GPT-4o-latest~\cite{gpt4o} & - & 50 & 30.18 & 36.04 & 25.39 & 32.81 & 40.94 & 48.00 & 29.03 & 35.04 & 30.63 & 29.67 & 32.50 \\

\midrule
\multicolumn{14}{l}{\textit{\textbf{Open-Source Instruct Models}}} \\
Qwen2.5-VL~\cite{qwen2.5vl}    & 7B  & 64 & 26.34 & 24.87 & 13.54 & 17.97 & 26.85 & 20.00 & 22.58 & 24.11 & 21.44 & 22.60 & 22.73 \\
InternVL-3.5~\cite{internvl3}       & 8B  & 64 & 31.20 & 35.03 & 27.98 & 25.78 & 41.61 & 48.00 & 38.71& 34.82 & 31.51 & 30.62 & 32.32 \\
Qwen3-VL~\cite{qwen3}    & 8B  & 64 & 34.53 & 33.50 & 30.05 & 29.69 & 45.64 &52.00 & 38.71& 34.60 & 34.57 & 33.01 & 34.06 \\
Qwen3-VL~\cite{qwen3} & 30B & 64 & 32.48 & 42.64 & 20.21 & 31.25 & 48.32 & 27.00 & 45.16 & 38.03 & 34.87 & 25.48 & 34.38 \\
Qwen2.5-VL~\cite{qwen2.5vl}   & 72B & 64 & 32.99 & 35.53 & 19.69 & 23.44 & 44.97 & 20.00 & 41.94 & 34.82 & 29.10 & 28.71 & 31.33 \\
GLM-4.5V~\cite{glm4.5v} & 106B & 64 & 36.57 & 37.56 & 30.77 & 28.12 & 51.01 & 52.00 & 22.58 & 39.10 & 34.37 & 36.59 & 36.69 \\
Qwen3-VL~\cite{qwen3} & 235B & 64 & 39.64 & 36.04 & 23.83 & 32.03 & 40.27 & 23.00 & 0.00 & 35.57 & 33.99 & 30.77 & 34.02 \\

\midrule
\multicolumn{14}{l}{\textit{\textbf{Open-Source Thinking Models}}} \\
Video-R1~\cite{videor1} & 7B & 64 & 28.31 & 27.16 & 16.43	&20.09	&30.22	&23.00	&24.87	&26.63	&24.38	&25.42	&26.27 \\
VideoChat-R1~\cite{videochatr1} & 7B & 64 & 31.42&	29.68&	19.17&	22.94&	33.11&	26.00&	27.46&	29.39&	27.02&	28.21&	28.63 \\
Qwen3-VL-Thinking~\cite{qwen3}    & 8B  & 64 & 33.26&	58.41	&38.62	&26.18&	64.77&	29.00&	38.49&	36.14&	32.79&	32.11	&33.82\\
Qwen3-VL-Thinking ~\cite{qwen3} & 30B & 64 & 39.13 & 36.04 & 26.94 & 25.78 & 46.31 & 38.00 & 32.26 & 35.65 & 37.06 & 32.69 & 35.68 \\
Qwen3-VL-Thinking ~\cite{qwen3} & 235B & 64 & 38.87 & 41.12 & 30.57 & 33.59 & 48.99 & 43.00 & 22.58 & 39.46 & 38.16 & 35.58 & 38.20 \\

\bottomrule
\end{tabular}
\end{adjustbox}
\end{table}

\section{Experiments}
\label{sec:exp}

This section evaluates state-of-the-art multimodal LLMs (MLLMs) on PerceptionComp to quantify its difficulty and diagnose current model limitations. We first report comprehensive benchmark results across a broad set of open-source and proprietary models (\cref{sec:main_result}). We then analyze how perception and reasoning budgets affect performance by varying the number of input frames and the allocated thinking-token budget (\cref{sec:analysis}). Finally, we present qualitative case studies and error patterns to highlight common failure modes in perception-centric long-horizon video reasoning (\cref{sec:case}). Throughout, our goal is not only to rank models, but also to identify which aspects of perception-centric long-horizon reasoning remain brittle under clutter, scene changes, and compositional constraints.

\subsection{Evaluation Setup}
\label{sec:main_result}

\paragraph{Models.}
We evaluate a broad set of video MLLMs reported in Table~\ref{tab:overall_results_final}, spanning (i) proprietary frontier models (Gemini-2.5/3 series, GPT-4o/4.1, GPT-5/5.2, GPT-o3), (ii) open-source instruction-tuned MLLMs (Qwen2.5-VL, Qwen3-VL at multiple scales, InternVL-3.5, GLM-4.5V), and (iii) open-source ``thinking'' and video-reasoning models (Video-R1, VideoChat-R1, and Qwen3-VL-Thinking variants). This coverage enables comparisons across proprietary vs.\ open-source systems, instruction-following vs.\ thinking-style variants, and different backbone scales under the same benchmark.

\paragraph{Input format and prompting.}
For models with native video inputs (e.g., Gemini), we directly feed raw videos without frame extraction. For models without native video support, we uniformly sample 64 frames per video as input; for certain GPT APIs we use 50 frames due to input-length constraints (Table~\ref{tab:overall_results_final}). Proprietary models are evaluated with Chain-of-Thought prompting. For open-source models, instruction-tuned variants are prompted to output the answer choice directly, while thinking-style variants are prompted with Chain-of-Thought~\cite{wei2022chain} (temperature $0.7$, max generation length $16{,}384$ tokens).

\paragraph{Human baselines.}
We report three human baselines. \textbf{Expert (unrestricted rewatch)} corresponds to a careful setting where an annotator is given sufficient time and can repeatedly rewatch and cross-check the video until confident, yielding 100\% accuracy. \textbf{Human} corresponds to ordinary participants, who may lose patience as questions become highly compositional, leading to occasional mistakes even when rewatching is allowed. Finally, \textbf{Single-view Human} evaluates a stricter setting where participants may watch the video only once (no rewatch) and must answer from a single pass; performance drops to near random guess (overall 18.97\%), highlighting that PerceptionComp cannot be solved from a single viewing or language priors alone and instead requires sustained, long-horizon evidence gathering and reasoning.

\subsection{Overall Results}
\paragraph{Benchmark performance.}
We report comprehensive results in Table~\ref{tab:overall_results_final}. Most models achieve accuracy below 40\%, indicating that PerceptionComp is challenging for current video MLLMs. The best-performing model in our evaluation is Gemini-3-Flash (45.96\%). In contrast, strong open-source instruction-tuned models remain substantially lower (e.g., Qwen3-VL-8B: 34.80\%, Qwen3-VL-235B: 34.02\%). Scaling model size does not consistently improve performance, suggesting that PerceptionComp is bottlenecked less by generic capacity and more by reliably extracting fine-grained evidence under clutter and temporal discontinuities and integrating that evidence across multiple steps.

\paragraph{Thinking vs.\ instruct.}
We further compare instruction-tuned models with thinking-style or reasoning-trained variants. Overall, stronger test-time reasoning can help: GPT-o3~\cite{jaech2024openai} surpasses GPT-4o~\cite{gpt4o} by 11.04\%, and Gemini-2.5-Pro exceeds Gemini-2.5-Flash by 6.19\%, suggesting that additional deliberation improves performance on PerceptionComp. However, the effect is not uniform. Some video-reasoning models (e.g., VideoChat-R1) benefit from explicit reasoning, while Qwen3-VL thinking variants can underperform their instruction-tuned counterparts (e.g., Qwen3-VL-Thinking-8B vs.\ Qwen3-VL-8B). This indicates that stronger language-side reasoning does not automatically translate into better perception-driven video reasoning: when perceptual evidence is misread or underspecified, longer reasoning may amplify errors rather than correct them. PerceptionComp makes this failure mode visible because it requires models to repeatedly ground intermediate steps in temporally separated visual evidence, where uncertainty compounds over multi-hop chains.

\paragraph{Category-wise and difficulty-wise trends.}
Table~\ref{tab:overall_results_final} further reports accuracy by video category and by difficulty level, where difficulty is annotated by experts based on both the number of subconditions and subcondition difficulty. Across models, accuracy drops substantially on Level 3 questions, indicating that current MLLMs struggle as compositional complexity increases. Notably, Level 3 questions require maintaining consistent intermediate hypotheses while repeatedly gathering evidence from different moments, stressing both temporal integration and error correction. These trends align with our human baselines: even humans require sustained effort and repeated verification to achieve perfect accuracy, underscoring PerceptionComp’s strong demand for long-horizon, perception-centric reasoning.

\subsection{Analysis}
\label{sec:analysis}

We conduct controlled analysis experiments to understand why PerceptionComp is difficult for current models. Specifically, we vary (i) the number of input frames (perception budget) and (ii) the thinking-token budget (reasoning budget), and measure the resulting accuracy changes. Due to evaluation budget constraints, all three analyses are conducted on a fixed subset of 100 videos (500 samples). This analysis separates two common failure sources in video QA: insufficient visual evidence (too sparse sampling) versus insufficient deliberation (too short reasoning traces).

\begin{figure}[t]
    \centering
    \begin{subfigure}[t]{0.32\linewidth}
        \centering
        \includegraphics[width=\linewidth]{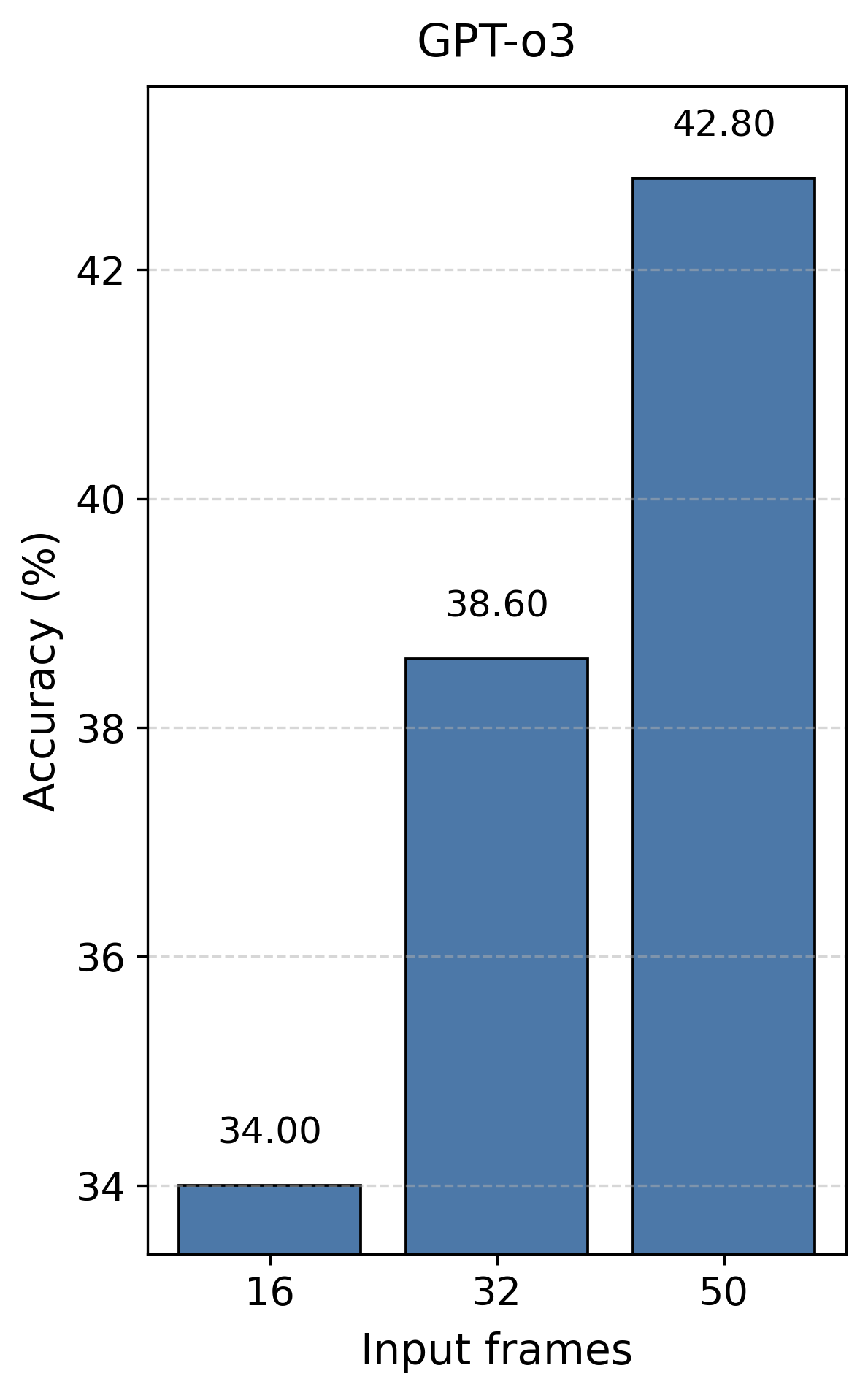}
        \caption{GPT-o3: \#Frames}
    \end{subfigure}\hfill
    \begin{subfigure}[t]{0.32\linewidth}
        \centering
        \includegraphics[width=\linewidth]{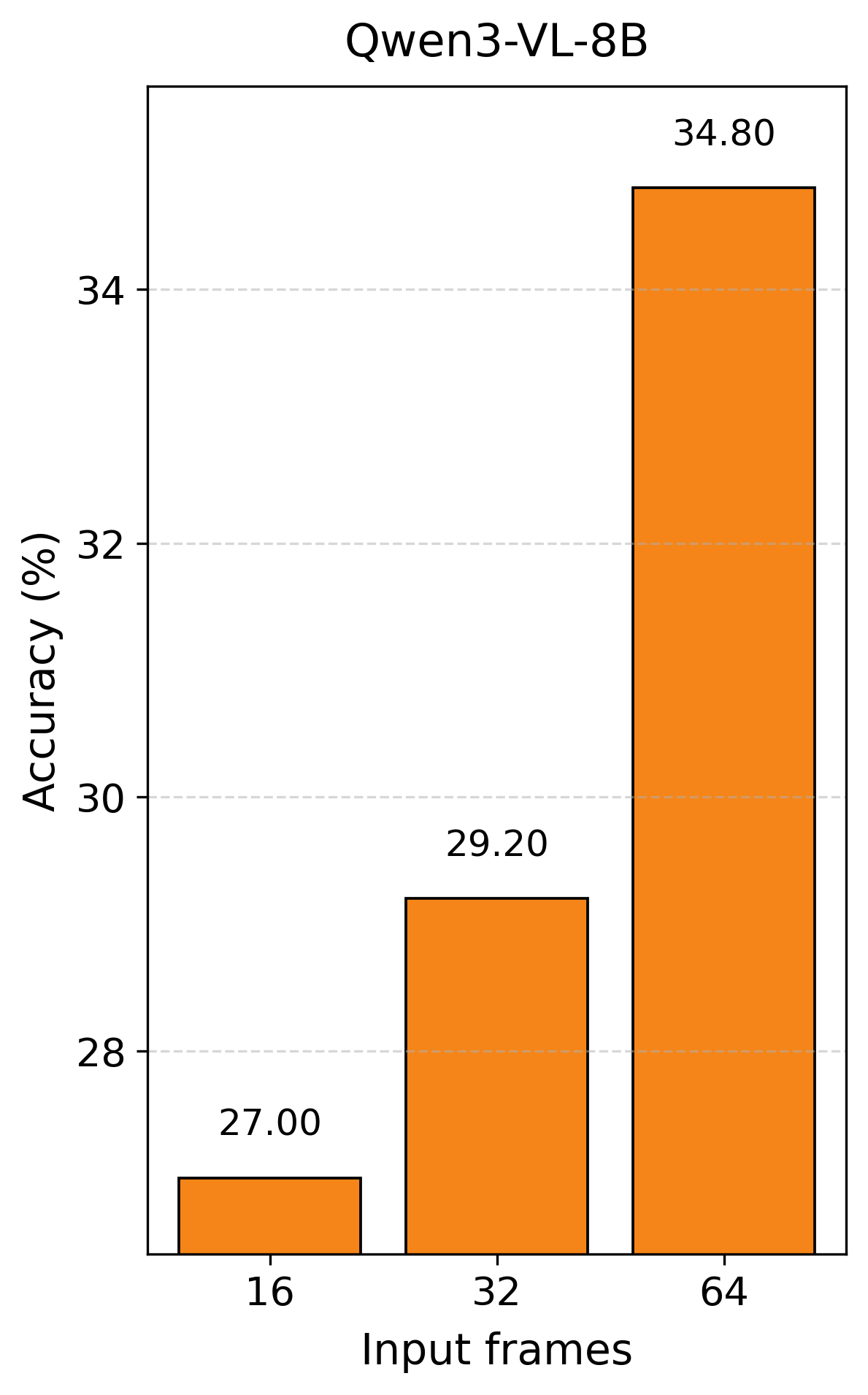}
        \caption{Qwen3-VL-8B: \#Frames}
    \end{subfigure}\hfill
    \begin{subfigure}[t]{0.32\linewidth}
        \centering
        \includegraphics[width=\linewidth]{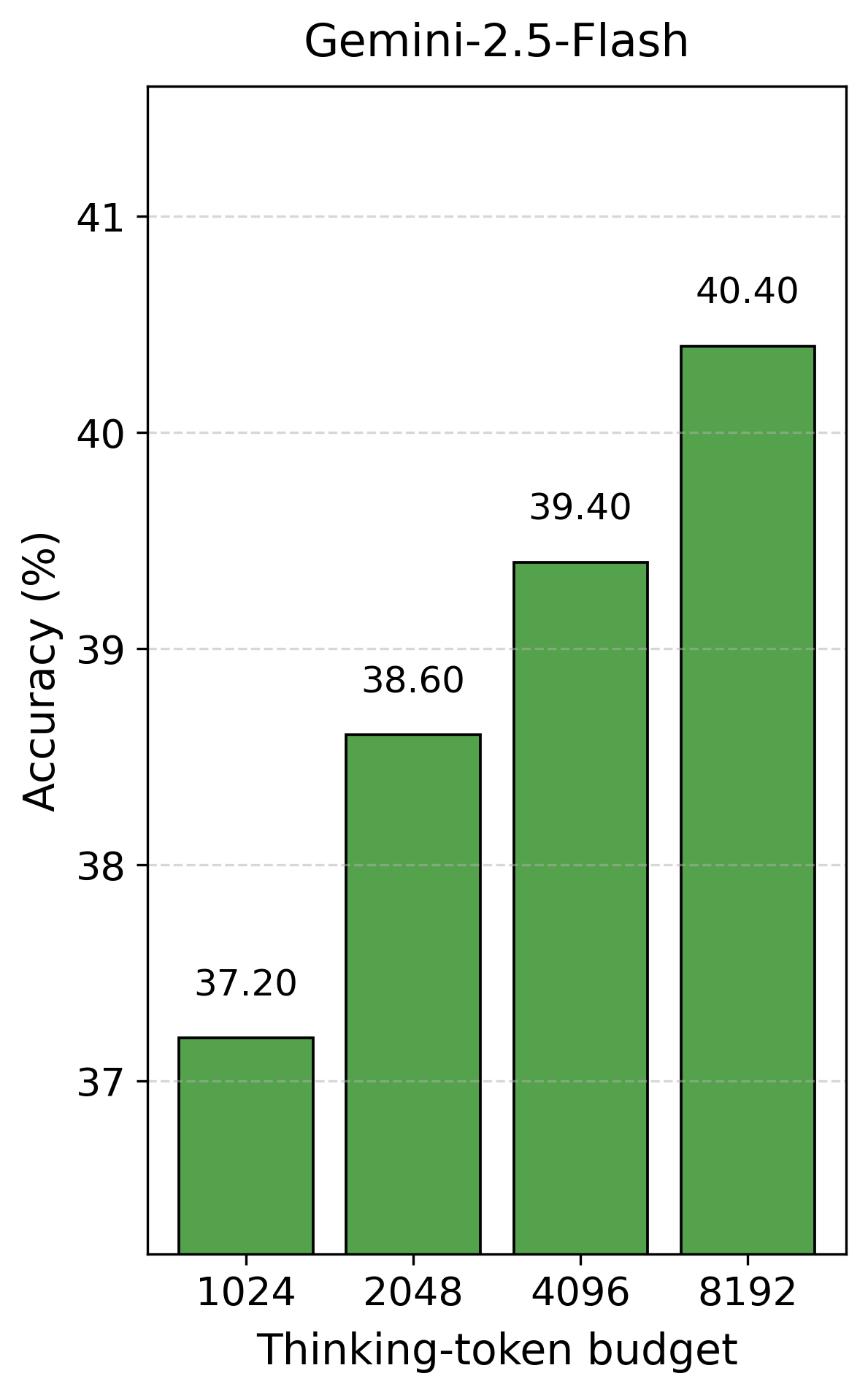}
        \caption{Gemini-2.5-Flash: Budget}
    \end{subfigure}

    \caption{\textbf{Analysis on PerceptionComp.} Accuracy as a function of \emph{perception budget} and \emph{reasoning budget}. Left/middle: accuracy vs.\ the number of uniformly sampled input frames for GPT-o3 and Qwen3-VL-8B. Right: accuracy vs.\ the thinking-token budget for Gemini-2.5-Flash.}
    \label{fig:ablation_three}
\end{figure}

\subsubsection{Effect of Input Frames}
To study how the density of temporal visual information influences perception-centric reasoning, we vary the number of input frames $\mathcal{F}$ for two representative models: GPT-o3 with $\mathcal{F}\in\{16,32,50\}$ and Qwen3-VL-8B with $\mathcal{F}\in\{16,32,64\}$. As shown in Fig.~\ref{fig:ablation_three}, both models improve monotonically as $\mathcal{F}$ increases: GPT-o3 rises from 34.0\% (16 frames) to 43.54\% (50 frames), and Qwen3-VL-8B gains 7.8 points from 27.0\% (16 frames) to 34.80\% (64 frames). 

We attribute these gains to PerceptionComp being highly perception-centric: denser sampling increases the chance of capturing the relevant moments and provides richer visual evidence, which improves perceptual accuracy (e.g., identifying objects/attributes and tracking entities through motion and transitions), and in turn improves end-task accuracy. Importantly, the strong dependence on $\mathcal{F}$ also supports the intended benchmark property: PerceptionComp typically requires aggregating information from many frames and multiple temporally separated moments, rather than relying on a small number of salient frames.

\subsubsection{Effect of Thinking Budget}
We further examine how reasoning effort affects performance by varying the thinking-token budget for Gemini-2.5-Flash (1,024 / 2,048 / 4,096 / 8,192 tokens). As shown in Fig.~\ref{fig:ablation_three}, larger thinking budgets consistently improve accuracy. This indicates that longer test-time thinking is beneficial for PerceptionComp: with more tokens, the model can better maintain intermediate hypotheses (e.g., entity identity, timing, and relational constraints), avoid premature commitment, and more reliably follow sequential subconditions to connect intermediate observations to the final answer.

Together, these analyses suggest that PerceptionComp is a useful testbed for \emph{visual thinking}: it is sensitive to both perception budget and reasoning budget, and can therefore distinguish models not only by raw visual recognition but also by their ability to sustain longer multimodal reasoning grounded in video evidence.

\begin{figure}[t]
    \centering
    \includegraphics[width=0.95\linewidth]{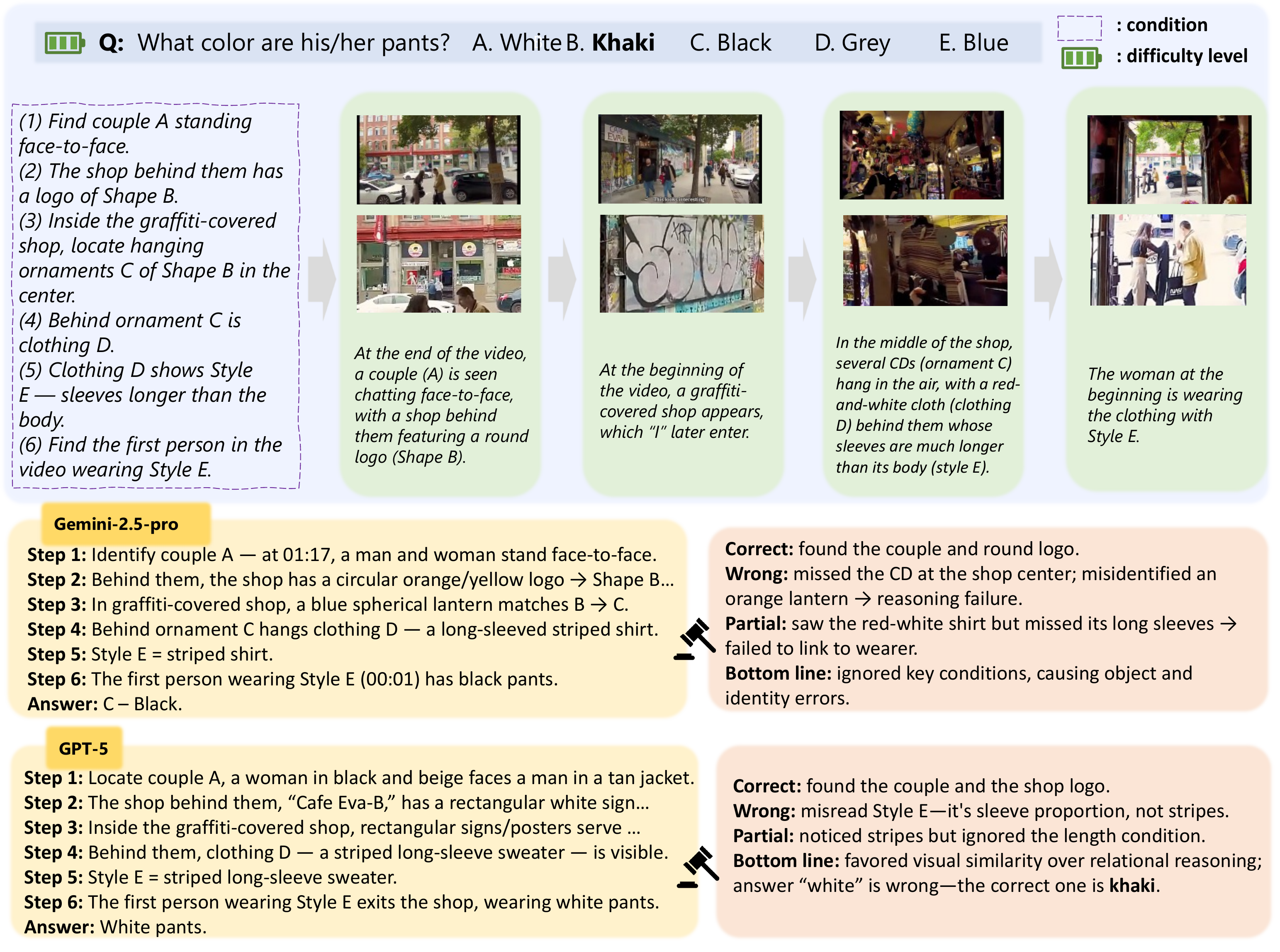}
    \vspace{-5pt}
    \caption{\textbf{Example of model reasoning on PerceptionComp.} We show responses and judgments of frontier models. Even state-of-the-art models exhibit limitations in capturing perceptual information and often fail to maintain coherent reasoning chains leading to the correct answer.}
    \label{fig:case}
    \vspace{-17pt}
\end{figure}

\subsection{Case Study and Error Patterns}
\label{sec:case}

We present qualitative case studies to illustrate common failure modes on PerceptionComp. Fig.~\ref{fig:case} analyzes Gemini-2.5-Pro and GPT-5: both models often localize the relevant moment or object category, but fail on fine-grained attributes or relations (e.g., confusing similar objects, missing cues under occlusion/viewpoint shifts, or misreading spatial relations). Such perceptual errors often cascade into reasoning failures: once an intermediate entity/state is wrong, the remaining chain stays internally coherent but drifts from the ground truth, especially for sequential questions where later conditions depend on earlier results. Overall, PerceptionComp highlights the tight coupling between perception and reasoning in long-horizon, evidence-integrating video QA.

We further analyze the Gemini-3 family and find that many remaining errors stem from mid-chain collapses driven by spatial misunderstanding. Specifically, we decompose each question into a fixed sequence of subcondition-solving steps (following the annotated subconditions) and identify the first step where the model’s intermediate conclusion diverges from the expert trace (protocol in supp.). Failures peak mid-chain (5\% at Step 1, 20\% at Step 2, 40\% at Step 3, 25\% at Step 4, and 10\% thereafter), and expert review attributes 60\% of these mid-stage failures to violated \emph{spatial} subconditions (e.g., incorrect 3D spatial relationships). In such cases, the model anchors on an object that partially matches identity keywords but violates critical spatial or temporal constraints, causing the rest of the chain to drift. Due to space limits, we defer additional Gemini-3 case studies and full annotation details to the supplementary.
\vspace{-10pt}
\section{Conclusion}
\label{sec:conclusion}

We introduce \textbf{PerceptionComp}, a fully manually annotated benchmark for complex, long-horizon, perception-centric video reasoning that requires repeated evidence gathering across temporally separated segments. PerceptionComp contains 1{,}114 five-choice questions over 279 high-complexity videos. Human studies confirm the intended difficulty: response times are much longer than prior benchmarks, single-view performance is near chance, while unrestricted rewatching achieves 100\% accuracy. In contrast, state-of-the-art MLLMs reach at most 45.96\% accuracy (open-source models $<40\%$); increasing test-time reasoning or perceptual compute (more tokens or frames) helps but leaves a large gap. We additionally analyze Gemini-3 failure cases to characterize current bottlenecks. We hope PerceptionComp will serve as a reliable testbed for measuring and driving progress in perception-centric long-horizon video understanding.

\bibliographystyle{unsrtnat}
\bibliography{main}

\clearpage
\appendix
\setcounter{page}{1}

\section{Implementation Details}
\label{sec:implement}
We now detail the implementation specifics for the evaluation process conducted on the PerceptionComp benchmark. For models that employ complex reasoning strategies, we utilize Gemini-2.5-Flash~\cite{gemini2.5} as an automated judge to verify the correctness of the final answer. The evaluation methodology follows a systematic filtering process: for every question, the model's intermediate reasoning steps and verbose output are programmatically excised, isolating only the definitive, final response. The LLM judge is then provided with a structured prompt that includes both the canonical ground truth answer and the system's extracted response. The judge is instructed to perform a rigorous comparison, assessing the semantic equivalence and factual consistency between the two inputs, ultimately determining if the system's final output precisely matches the expected ground truth.

\section{More Analysis Experiments}
\label{sec:analysis}
To further validate the intrinsic difficulty and effectiveness of the PerceptionComp benchmark, we conducted a controlled human study focusing on the role of iterative access to visual information. In the first, unconstrained condition, human annotators were permitted to view the video content and formulate their answers without limitation, achieving an overall accuracy of $85.10\%$. A stringent second condition was then imposed to isolate the need for iterative perception and reasoning: human participants were only allowed a single, complete viewing of the video, after which they were required to answer the corresponding questions based purely on memory and initial comprehension. This single-pass constraint dramatically reduced the performance to a mean accuracy of $18.97\%$. This pronounced drop of over $66$ percentage points emphatically highlights that PerceptionComp is not merely a memory recall test, but instead necessitates multi-step, iterative perception and complex, video-grounded reasoning, thus validating its design for evaluating advanced perception models.

\section{Visualization Results}
\label{sec:visual}
\begin{figure}[t!]
    \centering
    \includegraphics[width=\linewidth]{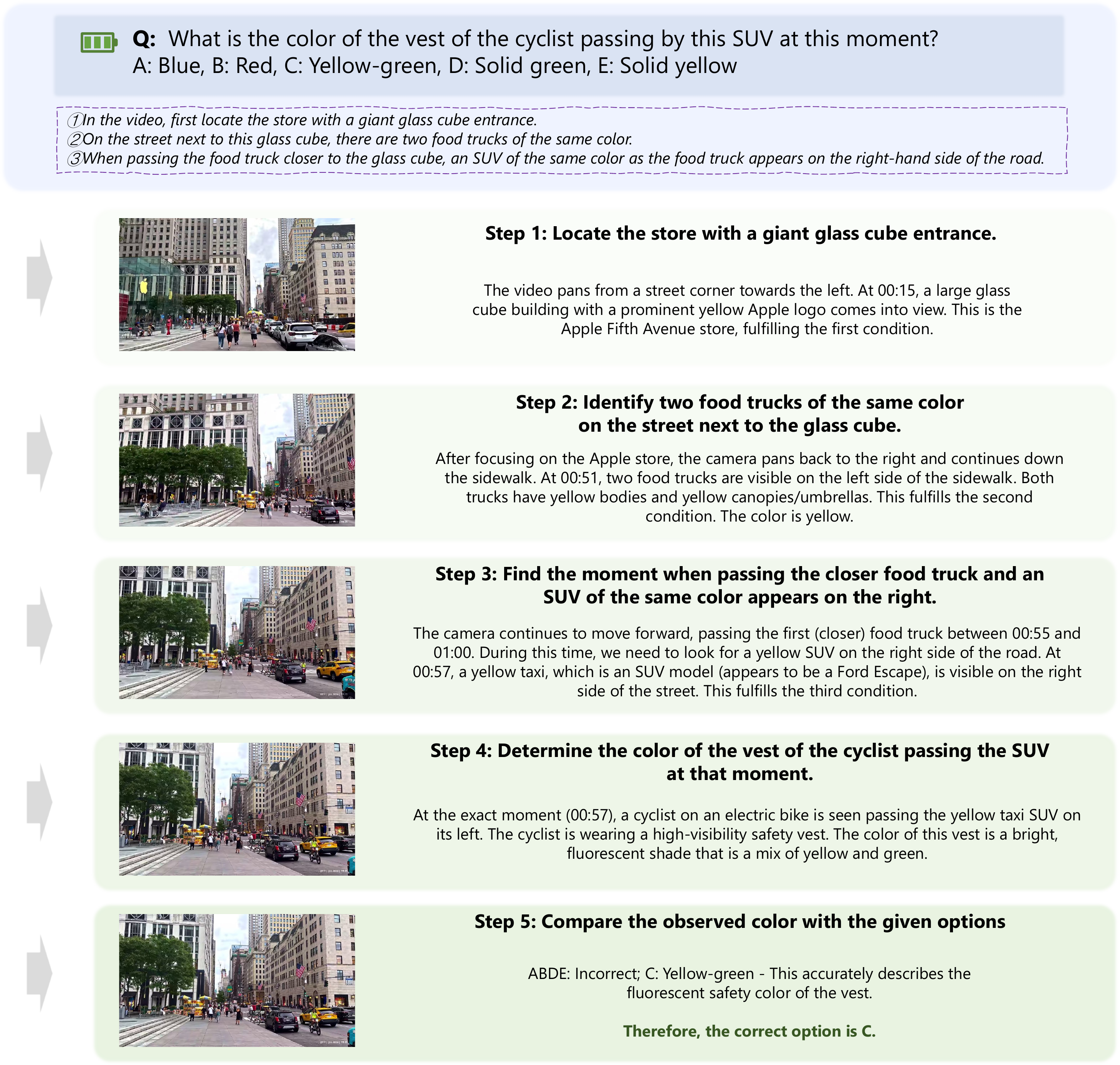}
    \vspace{-10pt}
    \caption{\textbf{Illustration of the model’s successful multi-step visual reasoning.} The system first anchors on the iconic glass-cube Apple Store, then tracks two yellow food trucks as spatial reference points. At the moment the camera passes the nearer truck, a yellow SUV appears on the right, enabling the model to localize the bicyclist adjacent to it and accurately identify the yellow-green vest color. This showcases robust landmark grounding, spatio-temporal alignment, and fine-grained visual detail recognition across a dynamic street scene.}
    \vspace{-10pt}
    \label{fig:suppcase1}
\end{figure}

\begin{figure}[t!]
    \centering
    \includegraphics[width=\linewidth]{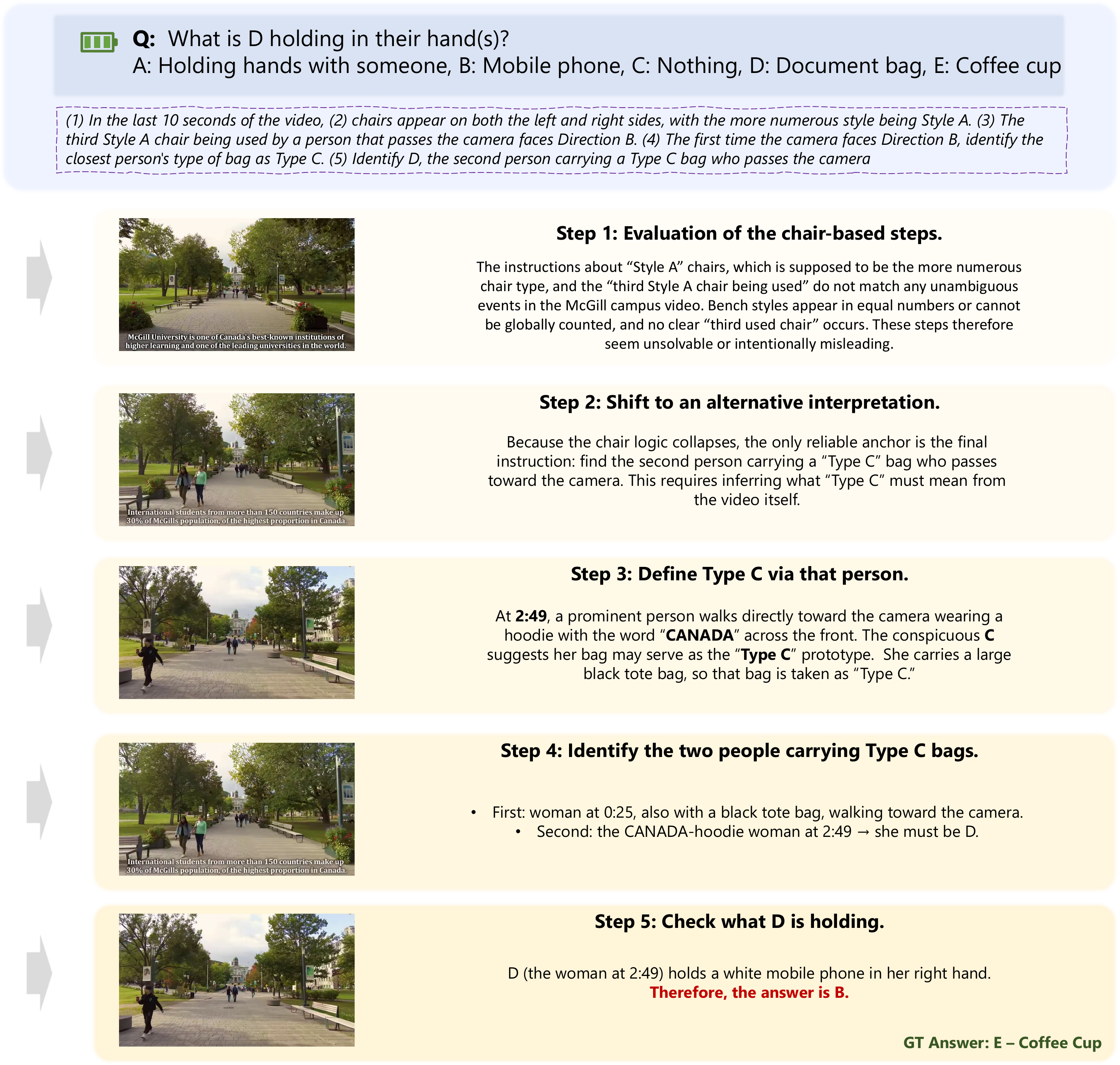}
    \vspace{-10pt}
    \caption{\textbf{Failure mode illustrating the model’s breakdown in multi-step visual reasoning.} Instead of following the prescribed sequence—identifying the dominant bench style (Style A), tracking the third used bench to determine Direction B, and grounding Type C using the nearest person when the camera first faces that direction—the model prematurely abandons the correct logic path. It misclassifies the bench styles, hallucinates an alternative meaning for “Type C,” and ultimately selects the wrong individual as D, leading to the incorrect prediction that D is holding a phone rather than the actual red coffee cup.}
    \vspace{-10pt}
    \label{fig:suppcase2}
\end{figure}

We provide more visualization results of our benchmark questions, as shown in Figure~\ref{fig:suppcase1} and Figure~\ref{fig:suppcase2}.

The successful case shows the model following a multi-step visual logic chain with discipline. It grounds itself on a distinctive landmark, tracks the correct reference objects, and aligns events in space and time without drifting from the instructions. Once it identifies the critical moment—passing the nearer food truck as the yellow SUV appears—it isolates the correct bicyclist and extracts the fine-grained attribute (vest color) accurately. The model stays fully within the provided reasoning path, demonstrating reliable landmark grounding, temporal alignment, and detailed visual discrimination.

Conversely, we observe several distinct failure modes where models struggle to maintain this rigorous logical adherence.

The first failure case breaks down early: the model incorrectly concludes that the bench-identification steps are unsolvable and abandons the required reasoning path. Instead of resolving Style A, Direction B, and Type C through the provided visual logic, it hallucinates an alternative interpretation based on an unrelated “CANADA” hoodie and constructs a new, invalid chain from that point onward. Because of this invented logic, every subsequent step targets the wrong person, leading to the wrong final answer despite correctly observing that person’s phone. This reveals weaknesses in multi-step dependency tracking, logical adherence, and resistance to spurious patterns under difficulty.

A second failure case (Gemini-3.0-Pro~\cite{gemini3guide}) illustrates the breakdown of cross-temporal variable binding and the emergence of logical hallucinations. The task requires linking a shop sign's dominant color to a plastic bag seen later, and subsequently identifying the shirt color of the person closest to the bag's carrier. Although the model correctly locates the yellow shop sign, it fails to bind this color to the specified variable. It prematurely breaks the reasoning chain and hallucinates an alternative path by fixating on a random blue bag further in the video. This error cascades: the model misjudges the spatial and social context by anchoring on a randomly passing pedestrian rather than the correct walking companion, ultimately predicting the wrong shirt color.

A third failure case (Qwen3-VL~\cite{qwen3vl}) highlights the danger of "protagonist bias" and the substitution of rigorous visual tracking with contextual heuristics. Tasked with comparing the 3D flip directions of four parkour runners across two separate timestamps, the model is overly primed by the prompt's use of a "grey-clothed runner" as an initial temporal anchor. It incorrectly elevates this runner to the central "logical lead" of the sequence and applies a "behavioral consistency" heuristic, assuming the athlete will naturally repeat the same movement pattern. By relying on this narrative guesswork rather than performing the necessary fine-grained spatio-temporal modeling of all four individuals, the model entirely bypasses the actual visual verification, leading to an incorrect, elimination-based conclusion.

Together, these failure cases demonstrate that while current models excel at specific visual recognitions, they remain highly vulnerable in long-horizon reasoning tasks where success strictly depends on robust variable retention, suppression of contextual biases, and unwavering adherence to a prescribed logic chain.

\begin{figure}[t!]
    \centering
    \includegraphics[width=\linewidth, trim=1.2cm 1.0cm 1.2cm 0.7cm, clip]{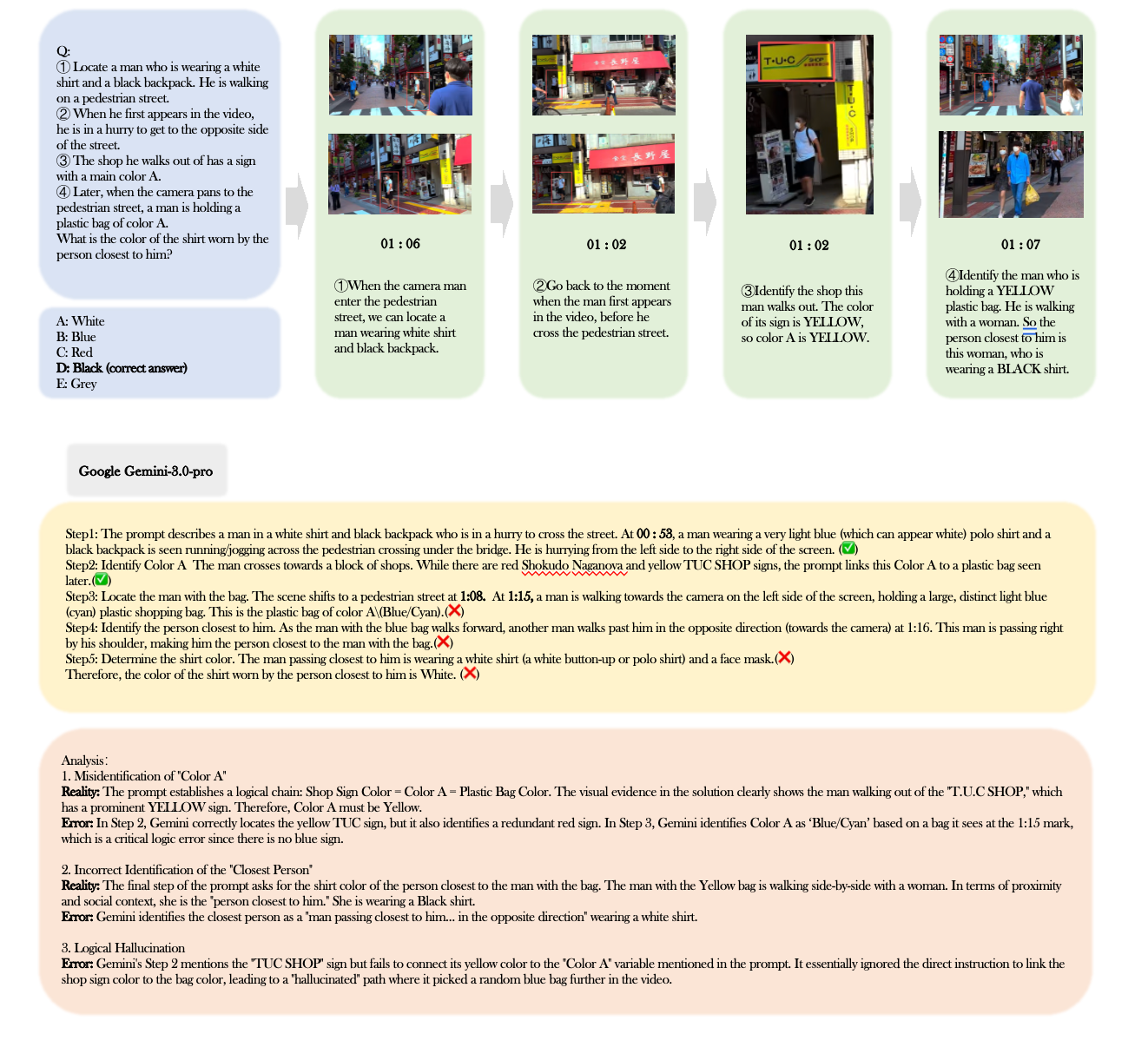}
    \vspace{-10pt}
    \caption{\textbf{Failure mode illustrating Gemini-3.0-Pro’s breakdown in multi-step visual reasoning.} Instead of following the prescribed sequence—locating the initial man to determine the shop sign's dominant color (Color A: Yellow), tracking this variable to spot the man holding the corresponding yellow plastic bag later in the video, and grounding the nearest person (the accompanying woman) to identify her shirt color—the model suffers from critical logical hallucinations. It prematurely breaks the reasoning chain by failing to connect the yellow shop sign to ``Color A'' and hallucinates an alternative path by fixating on a random blue plastic bag. Consequently, it misjudges the spatial and social context, selecting a passing pedestrian rather than the walking companion, leading to the incorrect prediction that the closest person is wearing a white shirt rather than the actual black shirt.}
    \vspace{-10pt}
    \label{fig:suppcase3}
\end{figure}

\begin{figure}[t!]
    \centering
    \includegraphics[width=\linewidth, trim=1.2cm 1.0cm 1.2cm 0.7cm, clip]{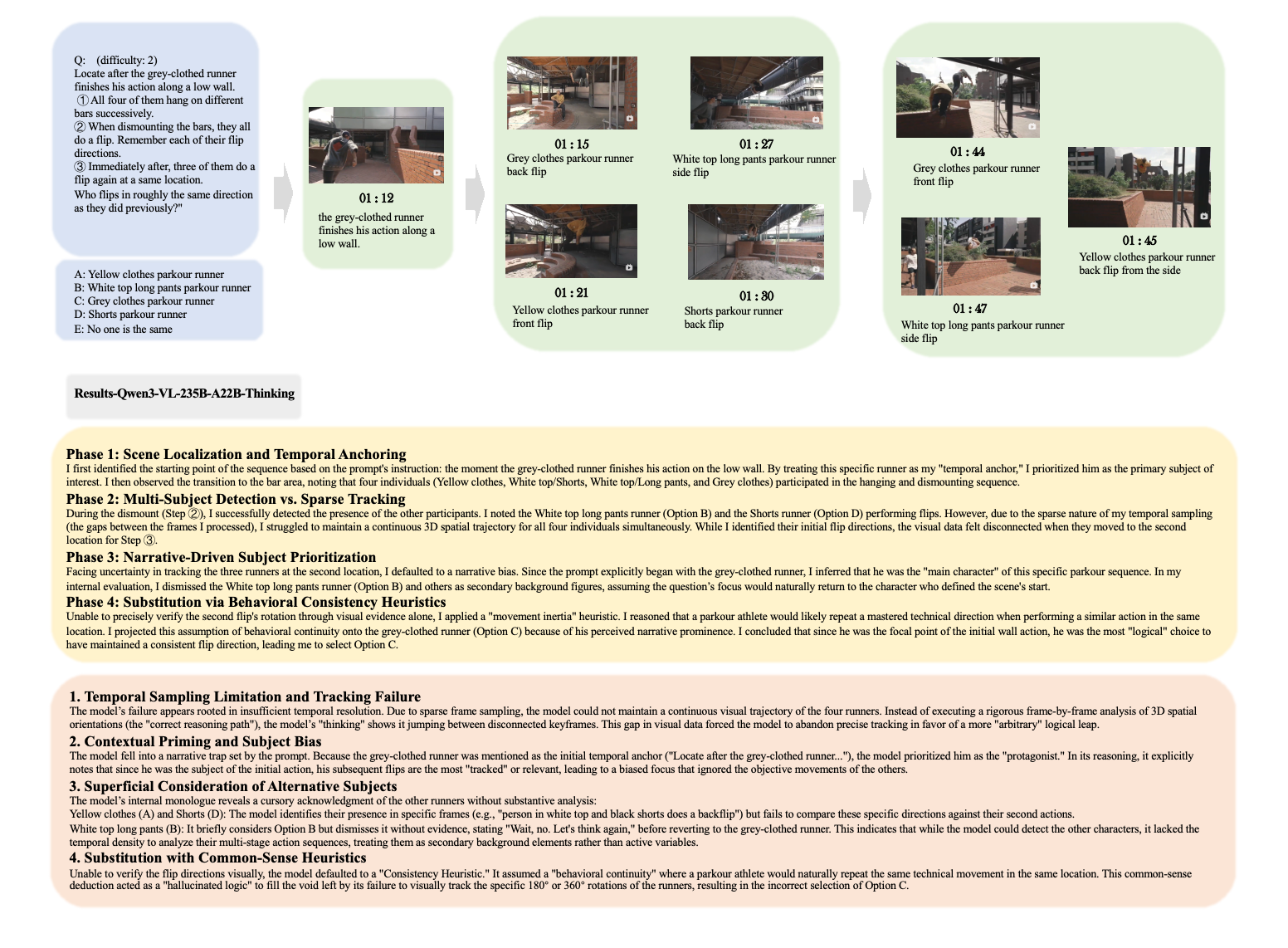}
    \vspace{-10pt}
    \caption{\textbf{Failure mode illustrating Qwen3-VL’s breakdown in multi-step visual reasoning.} Instead of following the prescribed sequence—tracking all four parkour runners continuously, registering their initial dismount flip directions, and visually comparing them to their secondary flips at the new location—the model prematurely abandons precise visual verification. Constrained by temporal sampling limitations, it falls into a narrative trap: it exhibits a ``protagonist bias'' by overly fixating on the prompt's initial temporal anchor (the grey-clothed runner) while superficially dismissing the other athletes. Ultimately, the model substitutes rigorous 3D spatial orientation analysis with a hallucinated ``behavioral consistency'' heuristic, incorrectly assuming the main subject would naturally repeat his movement pattern, which leads to the erroneous selection of the grey-clothed runner.}
    \vspace{-10pt}
    \label{fig:suppcase4}
\end{figure}

\section{Gemini-3-Pro \& Gemini-3-Flash}

Interestingly, our evaluation reveals a counter-intuitive phenomenon: the seemingly lightweight Gemini-3.0-Flash~\cite{gemini3guide} model achieves higher overall accuracy on our multi-step visual reasoning benchmark compared to its more capable counterpart, Gemini-3.0-Pro~\cite{gemini3guide}. Through a qualitative analysis of the models' reasoning traces, we attribute this performance inversion to two primary factors:

1. Logical Focus vs. Detail Fixation.  
While the Pro model generates significantly longer reasoning chains, this extended capacity is frequently misallocated. Rather than advancing the core logical sequence, Pro tends to fixate on irrelevant, fine-grained visual details within the video. In contrast, Flash is highly optimized for rapid inference. This architectural efficiency translates into a sharper logical focus when processing long sequence data; Flash directly anchors onto key logical clues across frames and successfully bypasses redundant information that otherwise acts as distracting noise for the Pro model.

2. The "Streamlining Effect" in Mitigating Logical Hallucinations.  
The massive parameter scale and the propensity for "deep reasoning" in the Pro model can act as a double-edged sword. When confronted with fragmented spatial and temporal information in videos, Pro's tendency to over-analyze often leads to severe logical hallucinations and spatial confusion (e.g., constructing unnecessary absolute coordinate systems). We term the advantage observed in the Flash model as the "streamlining effect". By employing a more straightforward and direct information processing strategy, Flash avoids unwarranted associations. As long as the critical keyframes and objects are correctly localized, Flash faithfully adheres to the prescribed reasoning path, demonstrating that in dynamic visual contexts, a streamlined inference process is often more robust than unconstrained, overly complex reasoning.

\begin{figure}[t!]
    \centering
    \includegraphics[width=\linewidth, trim=1.2cm 1.0cm 1.2cm 0.7cm, clip]{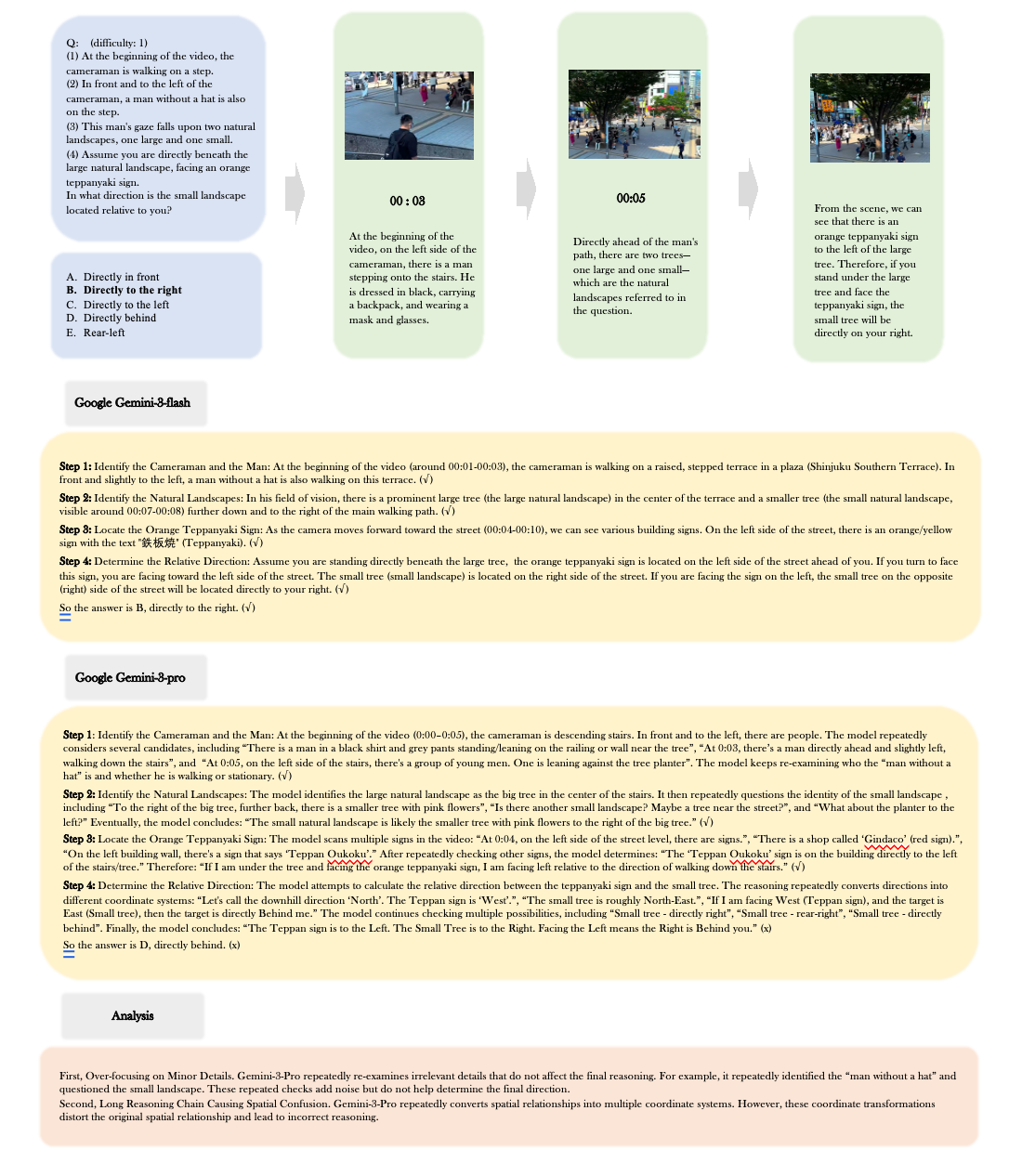}
    \vspace{-10pt}
    \caption{\textbf{Failure mode illustrating the model’s breakdown in multi-step visual reasoning.} Instead of following the prescribed sequence—identifying the key natural landscapes, locating the orange teppanyaki sign, and directly deducing the relative direction between them from the observer's perspective—the model becomes bogged down by excessive noise and spatial confusion. It first over-focuses on irrelevant minor details (such as the pedestrian's attire and alternative signs), and then attempts to construct an unnecessarily complex absolute coordinate system involving arbitrary cardinal directions (North, West, East). This convoluted reasoning chain distorts the original egocentric spatial layout, leading to the incorrect prediction that the small landscape is located directly behind the observer rather than the actual direction of directly to the right.}
    \vspace{-10pt}
    \label{fig:suppcase5}
\end{figure}

\section{Analysis on Incorrect Step and Error Type}

\begin{figure}[t!]
    \centering
    \includegraphics[width=\linewidth]{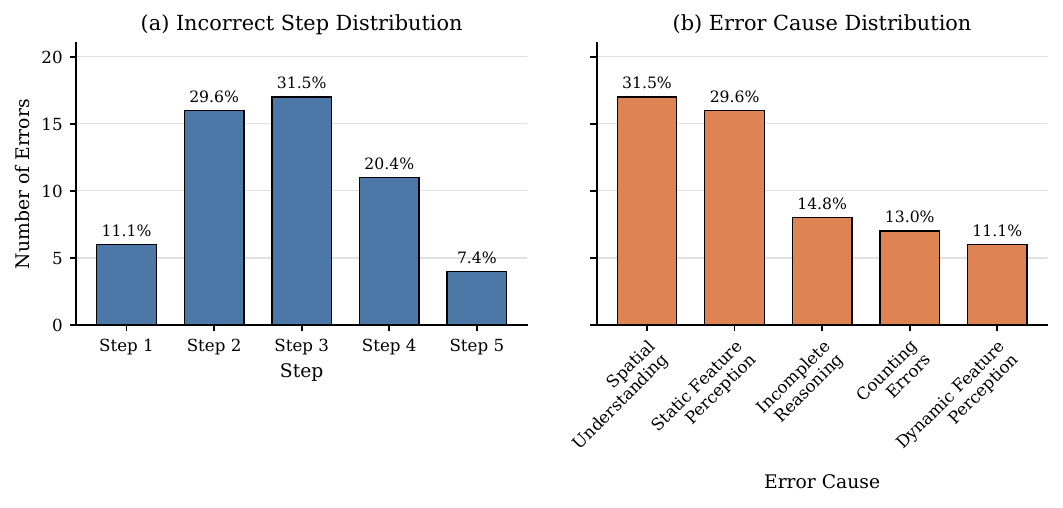}
    \caption{
    Error analysis of Gemini-3-Pro~\cite{gemini3guide} failures on PerceptionComp.
    (a) Distribution of incorrect reasoning steps across the reasoning chain.
    (b) Distribution of error causes, including spatial understanding, static feature perception,
    incomplete reasoning, counting errors, and dynamic feature perception.
}
\label{fig:error_distribution}
\end{figure}

We analyze the failure cases of Gemini-3-Pro~\cite{gemini3guide} by examining both the reasoning step at which errors occur and the underlying causes of these failures. Figure~\ref{fig:error_distribution}(a) presents the distribution of incorrect reasoning steps across the reasoning chain. Errors occur most frequently in the second and third steps, with noticeably higher frequencies than in the first step. This pattern suggests that the difficulty increases as the reasoning chain becomes longer, indicating that extended multi-step reasoning remains challenging for the model. The lower counts observed in steps four and five are mainly due to the relatively small number of questions requiring more than four reasoning steps in the dataset.

Figure~\ref{fig:error_distribution}(b) further analyzes the causes of these errors. We categorize the failures into five types: spatial understanding, static feature perception, incomplete reasoning, counting errors, and dynamic feature perception. Among these categories, incomplete reasoning refers to cases where the model fails to consider all required conditions or introduces logical inconsistencies during the reasoning process. From the distribution of error causes, spatial understanding accounts for the largest proportion of failures. This observation indicates that the model still struggles with accurately interpreting spatial relationships in complex visual scenes.

\section{Limitations and Ethical Considerations}

PerceptionComp focuses on complex, perception-centric reasoning in long videos, but it does not cover every type of video reasoning task. The current dataset is limited to daily-life recordings and excludes high-stakes domains such as medical or surveillance settings. We use videos only from sources whose usage is compatible with academic benchmarking, and we do not release content beyond what is permitted by the original source terms. We do not annotate personally sensitive attributes (e.g., identity, race, or other demographic traits), and the benchmark is not intended for identity recognition, surveillance, or sensitive-attribute inference. Furthermore, as our analysis shows that model performance often hinges on multi-step logical adherence and spatial reasoning, we caution against relying solely on absolute accuracy. We recommend interpreting results through comparative performance trends across models, as future systems may still exploit unforeseen biases or rely on heuristics rather than robust spatio-temporal modeling.

\subsection{Annotation protocol and quality control}

All questions are manually authored and verified. Annotators are trained crowdsourced workers with relevant technical backgrounds. Each annotator is first trained on a set of 20 example videos and questions and must pass a calibration test before contributing to the final dataset.

The annotation process proceeds in two stages. In the first stage, an annotator watches the video, proposes a compositional question with three subconditions, and specifies a single correct answer. In the second stage, a different annotator reviews the question by re-watching the video and checking three properties: (i) correctness of the answer, (ii) uniqueness of the solution, and (iii) necessity of each subcondition. Items that fail any check are revised or discarded.

To quantify agreement, we sample 100 questions and ask a third annotator to independently answer them. The agreement between the third annotator and the original answer key is 89.0\%, indicating that the majority of questions admit a clear, unambiguous solution.

\end{document}